\RequirePackage{fix-cm}
\documentclass[twocolumn, natbib]{svjour3}
\smartqed

\newif\ifanon\ifdefined\anonsubmission\anontrue\fi

\usepackage[utf8]{inputenc}
\usepackage[T1]{fontenc}
\usepackage{amsmath,amssymb}
\usepackage{booktabs}
\usepackage{array}
\usepackage{graphicx}
\graphicspath{{../../../figures/}}
\usepackage{textcomp}
\usepackage{xspace}
\usepackage{microtype}
\usepackage{needspace}
\usepackage[hidelinks]{hyperref}

\newcommand{\dg}{\ensuremath{^\circ}\xspace}

\journalname{PFG -- Journal of Photogrammetry, Remote Sensing and Geoinformation Science}

\begin{document}

\title{Hold-Out Self-Validation Cannot Certify the Accuracy of Photogrammetric
Reconstructions}
\titlerunning{Hold-out self-validation and photogrammetric accuracy}

\ifanon
\author{Anonymous Author(s)}
\authorrunning{Anonymous}
\institute{Withheld for anonymous review}
\else
\author{Behnam Asadi}
\authorrunning{B. Asadi}
\institute{B. Asadi \at
              Independent Researcher, Lausanne, Switzerland \\
              ORCID: 0009-0003-9246-7061 \\
              \email{behnam.asadi@gmail.com}
}
\fi

\date{Received: date / Accepted: date}

\maketitle

\begin{abstract}
Photogrammetric reconstructions used for inspection are often delivered without an
independent accuracy check. Hold-out
self-validation is a possible substitute. Images are withheld, their orientations are re-estimated from the rest of the block, and agreement is taken as evidence of quality.
We test it on ten GNSS-referenced aerial captures from eight sites, on ETH3D and EuRoC,
and on 30 IMC 2025 scenes. The standard form re-localises each
withheld image only against points seen by at least two retained images. Its score
saturates. It stayed at 1.00 in all 40 thinned reconstructions while camera-position error varied by a factor of up to 5.9 within one capture. It also scored 1.00 on a corrupted model with
only one of 226 cameras within 5\,m. Reprojection error ranked some of the worst models best.
A few lost cameras can raise the least-squares RMSE to the size of the block, so we also evaluate with a robust alignment.
Taking the correspondences from the verified matches lowered the score below 0.95 in 14 of 18 corrupted runs, against 4 for the standard form and 10 when its failed resections count as errors, with all seven nominal runs at 1.00.
Mapping the block again without the withheld images also responds, but mapping the same corrupted
data twice without withholding gave a similar disagreement, so the signal is
mainly that a corrupted reconstruction is not reproducible. Neither test can see an error
that every reconstruction repeats, and neither can replace check points.

\end{abstract}

\section{Introduction}\label{sec:intro}
Structure from motion (SfM) and multi-view stereo are now routine tools for inspecting
infrastructure. Facades, bridges, mine shafts and industrial sites are reconstructed
from drone or handheld images, and the results are used for measurement: the width of a
crack, the position of a point in a national grid. The accuracy of these numbers is
usually established with surveyed ground control and check points. Control is
expensive, sometimes impossible to place, and often left out of routine jobs. Many
reconstructions are therefore delivered with no independent accuracy statement at all.

Without check points, the quality of an image block can be judged only from quantities that
the adjustment itself reports: residuals, the reprojection error and the precision
derived from the covariance matrix. Classical reliability theory states what such internal
quantities can reveal. They detect only blunders that leave residuals in the adjustment
\citep{baarda1968,forstner1987reliability}. A block that is deformed as a whole but remains locally consistent leaves few such residuals, because the image observations constrain such a deformation only weakly, in particular when it is correlated with the interior orientation \citep{james2014mitigating}.

Another internal check is borrowed from cross-validation in machine learning. Some images are
withheld, the exterior orientation of each is estimated again from the rest of the
block, and the result is compared with the adjusted orientation. Applied directly to SfM,
this test carries little information. Each image helped to triangulate the points it observes, so
re-localising it against those points largely returns its own contribution. A common remedy, a track-length threshold, re-localises each withheld image only against points observed by at least two retained images. These points are still adjusted
together with the withheld image, so the threshold weakens the dependence but does not
remove it. A hold-out free of this dependence must also reconstruct the block without the withheld
images, which costs a full mapping run.

Reliability theory suggests that such a check will miss a deformation
that the block absorbs consistently. It does not say how large the undetected errors become in
practice, or whether the usual internal
indicators do any better. We ask two practical questions.
Does the hold-out score follow absolute accuracy closely enough to replace check points? If not, does it at
least detect gross failures? The protocol uses only the images
and the reconstruction. Our evaluation of it uses RTK positions, laser scans and benchmark
ground truth.

The contributions are as follows.
\begin{enumerate}
\item A statement of the hold-out protocol with a track-length threshold, and a worked example that shows what the threshold changes and which errors no hold-out test can detect.
\item Evidence from GNSS-referenced aerial captures and from ETH3D, EuRoC and IMC 2025
that the score saturates while absolute error varies widely.
\item A match-corruption experiment, evaluated with least-squares and robust alignment, showing that the standard score and
reprojection error miss models with many lost cameras and models that are wrong throughout. A variant with correspondences from verified matches and a
re-mapped hold-out respond better, mainly because the
corrupted solutions do not reproduce.
\end{enumerate}

\section{Related work}\label{sec:related}

\subsection{Reliability and datum definition}
Photogrammetry has long distinguished internal from external reliability
\citep{baarda1968,forstner1987reliability}. Internal reliability describes which gross
errors a redundant network can detect from its own residuals, and external reliability
how the undetected errors affect the parameters. In a free network the seven parameters of a similarity remain free \citep{triggs2000ba},
an ambiguity that changes no residual. A deformation of the block, such as
bending or a drifting scale, does change the observations, but
in a weak network or under a wrong camera model it may change them so little that the
adjustment absorbs it. \citet{james2014mitigating} showed that small errors in radial
distortion produce broad doming of UAV surveys in this way, and \citet{luhmann2016sensor}
discuss why high internal precision does not imply external accuracy.
Wrong correspondences can also produce a model that is internally consistent but
false. Criteria for
model selection and degeneracy, such as GRIC \citep{torr1997gric} and QDEGSAC
\citep{frahm2006qdegsac}, detect poorly conditioned configurations, not a well-conditioned model that is wrong.

\subsection{Quality assessment and uncertainty}
Benchmarks for multi-view reconstruction measure accuracy and completeness against laser
scans or reference meshes \citep{eth3d,tanks,dtu}. Pose benchmarks summarise rotation and
translation errors as mean average accuracy (mAA) over several angular thresholds
\citep{imc}. In aerial photogrammetry, accuracy is assessed against ground control,
independent check points and reference point clouds
\citep{nex2015benchmark,usegeo2024,h3d2021}. All of these need ground truth.

Covariance from the adjustment and learned uncertainty estimates describe precision, that
is, the spread of an estimate under noise \citep{kendall2017uncertainties}.
\citet{james2017precision} map this precision over SfM surveys to plan ground control.
Precision does not describe correctness when the functional model itself is wrong. \citet{lookma2024} tune SfM and SLAM
pipelines without ground truth by perturbing the inputs and measuring how the output
changes. Like our score, this signal does not respond to a deformation that the whole model shares.

\subsection{Re-localisation and hold-out validation}
Each step of our protocol resects a withheld image against an existing 3D model. This is
visual re-localisation, which has its own benchmarks \citep{sattler2018benchmarking} and methods \citep{sattler2017largescale,sarlin2019hfnet}. Re-localisation seeks the best pose against a
map that is assumed correct, whereas we use the disagreement between re-localised and
reconstructed poses to question the map itself. 

\section{Method}\label{sec:method}

\begin{figure*}[t]
\centering
\includegraphics[width=\textwidth]{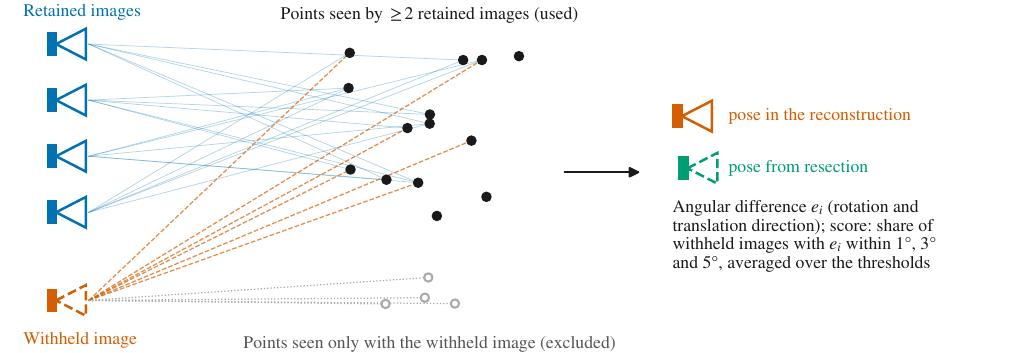}
\caption{Hold-out protocol. A withheld image is resected only against points seen by at least two retained images, and the score is computed from the angular difference between the resected and the reconstructed pose.}
\label{fig:method}
\end{figure*}

\subsection{Hold-out protocol}\label{sec:protocol}
Let an SfM reconstruction contain registered images $\mathcal{I}$ with poses
$(\mathbf{R}_i,\mathbf{t}_i)$, which map a world point to camera coordinates as
$\mathbf{x}=\mathbf{R}_i\mathbf{X}+\mathbf{t}_i$, and 3D points $\mathbf{X}_p$. Only the
largest connected model is scored. Each point has a track
$\mathcal{T}(p)\subseteq\mathcal{I}$, the set of images that observe it. The protocol has
four steps (Fig.~\ref{fig:method}).
\begin{enumerate}
\item Partition. A fraction of the images, 10\,\% by default, forms the withheld
set $\mathcal{H}$. The selection is a seeded random sample of the sorted image names, so the split is
reproducible for a given set of registered images.
\item Trusted points. The trusted set
$\mathcal{P}^\star=\{p : |\mathcal{T}(p)\setminus\mathcal{H}| \ge \tau\}$ with $\tau=2$
contains the points observed by at least two retained images.
\item Resection. For each $i\in\mathcal{H}$, the observations that the model stores for
image $i$ and that belong to points in $\mathcal{P}^\star$ serve as 2D--3D correspondences. They give a new pose $(\hat{\mathbf{R}}_i,\hat{\mathbf{t}}_i)$. We use a
P3P solver \citep{kneip2011p3p} inside RANSAC with a 12-pixel inlier threshold and
non-linear refinement of the pose, with the intrinsics of the full model held fixed.
Images with fewer than four correspondences or without a RANSAC solution form the set
$\mathcal{H}\setminus\mathcal{L}$ of failed resections.
\item Score. The rotation error $e^R_i$ is the geodesic angle between $\mathbf{R}_i$ and
$\hat{\mathbf{R}}_i$, and the translation error $e^t_i$ is the angle between the vectors
$\mathbf{t}_i$ and $\hat{\mathbf{t}}_i$. With $e_i=\max(e^R_i,e^t_i)$, the hold-out score
is
\begin{equation}
s=\frac{1}{|\Theta|}\sum_{\theta\in\Theta}\frac{1}{|\mathcal{L}|}
\sum_{i\in\mathcal{L}}\mathbb{1}[e_i\le\theta]
\label{eq:score}
\end{equation}
with $\Theta=\{1\dg,3\dg,5\dg\}$. The score is conditional on a successful resection, so
we report the coverage $c=|\mathcal{L}|/|\mathcal{H}|$ beside it. Counting failed resections as errors gives the product $c\,s$.
\end{enumerate}

Step 2 is a track-length threshold. Without it, a withheld image is resected against
points it helped to triangulate, and the resected pose largely reproduces the stored one.
The threshold removes points that depend on the withheld image alone or with one retained
image. The trusted points were still adjusted together with the withheld image, so the
test is a consistency check, not an independent prediction. The mapper also assigned these correspondences while estimating the pose of this image, so correspondences it rejected as outliers are excluded from the test.
Section~\ref{sec:remap} removes these dependences in two steps, by taking the correspondences from verified matches into
retained images, and by mapping the block again without $\mathcal{H}$.

The angle between $\mathbf{t}_i$ and $\hat{\mathbf{t}}_i$ is free of scale but depends on
the origin of the world frame, because $\mathbf{t}_i=-\mathbf{R}_i\mathbf{C}_i$ for the
projection centre $\mathbf{C}_i$. We kept it for comparability with pose benchmarks
\citep{imc}. Moving the origin of the 16 models of the experiment of Table~\ref{tab:failure} to the centroid of the cameras, or ten block extents away, changed
the score of 2 models, by at most 0.03.

\subsection{Worked example}\label{sec:example}
Table~\ref{tab:example} illustrates the dependence on a minimal scene with six points, four
retained cameras $C_1$ to $C_4$ and one withheld image $H$. In case (a), the stored
orientation of $H$ is rotated by 4.00\dg\ about its optical axis. All cameras look at the origin, so this roll leaves the direction of $\mathbf{t}_H$ unchanged. $P_6$ is seen by $H$ alone and is used only in the limiting row. If the trusted points $P_1$ to $P_4$ carried no contribution of $H$, the resection would show the full 4.00\dg\
error. Triangulated with the wrong ray of $H$ included, as in a real adjustment, the five points $P_1$ to $P_5$ hide about half of it, and the error falls to 2.10\dg. Points
lying exactly on the stored rays of $H$ hide all of it. In a real adjustment, the threshold moves the result from the third row of the table towards the second, but not to the first. In case (b), the whole scene is
scaled by 6\,\%. This is a similarity transformation and therefore lies within the datum
freedom. Both errors are exactly zero, although the points move by 0.072\,m RMS. A shear,
in contrast, changes the image residuals of calibrated cameras. 

\begin{table}[t]
\caption{Worked example with a rotation error of $H$ (a) and a 6\,\% scaling of the whole
scene (b), from P3P and EPnP resections \citep{kneip2011p3p,lepetit2009epnp}.}
\label{tab:example}
\centering
\small
\setlength{\tabcolsep}{4pt}
\begin{tabular}{@{}llcc@{}}
\toprule
Point & Track $\mathcal{T}(p)$ & $|\mathcal{T}(p)\setminus\mathcal{H}|$ & Trusted \\
\midrule
$P_1$ to $P_4$ & $\{C_i,C_j,H\}$ & 2 & yes \\
$P_5$ & $\{C_4,H\}$ & 1 & no \\
$P_6$ & $\{H\}$ & 0 & no \\
\bottomrule
\end{tabular}

\medskip
\begin{tabular}{@{}llccc@{}}
\toprule
Case & Points used & $e^R_H$ & $e^t_H$ & Correct at \\
\midrule
(a) & $P_1$--$P_4$, free of $H$ & 4.00\dg & 0.00\dg & 5\dg \\
(a) & $P_1$--$P_5$, with $H$ & 2.10\dg & 0.05\dg & 3\dg, 5\dg \\
(a) & $P_1$--$P_6$, on rays of $H$ & 0.00\dg & 0.00\dg & all \\
(b) & $P_1$--$P_4$ & 0.00\dg & 0.00\dg & all \\
\bottomrule
\end{tabular}
\end{table}

\subsection{Scope of the score}\label{sec:measures}
The score measures internal consistency, whether the pose of a withheld image
can be predicted from the rest of the model, not absolute accuracy.
An exact similarity of the whole model changes neither the score nor our evaluation, which removes it (Section~\ref{sec:reference}). The large errors reported below are of a different kind: lost cameras, damaged parts of the block and models that are wrong throughout, which remain after the best similarity has been removed. Our hypothesis is that a withheld image still fits its neighbourhood in such a model, so indicators computed from local consistency, such as reprojection
error, should miss the same errors (Section~\ref{sec:proxies}).

The translation error uses only the direction of the translation. Section~\ref{sec:sens} tests a variant that uses the
distance between the projection centres. It is as saturated as the original, so
the choice of error measure does not explain the saturation.

\section{Data and experiments}\label{sec:data}

\subsection{Datasets}\label{sec:datasets}
Ten aerial captures from eight sites have a GNSS
position for every image (Table~\ref{tab:data}). Four of them form the core of the accuracy analysis:
Helenenschacht, the two Tuniu captures and Bellus. The Tuniu captures show the same river
on two dates, so the core covers three sites. All four are public. Helenenschacht
\citep{odmhelenenschacht} and Bellus \citep{odmbellus} are OpenDroneMap sample datasets,
and the Tuniu captures were shared on the OpenDroneMap forum \citep{tuniu2019}. GUT Campus \citep{gutcampus} is an
independent fifth capture for the sparsity experiment. We used its 612 nadir images out
of a five-direction survey. The data also include three public surveys of the Baltic coast. Sellin \citep{sellin2025} and Wustrow
\citep{wustrow2024} enter the sensitivity tests and corruption experiments named in
Table~\ref{tab:data}, and a survey at Niehagen on the Dar{\ss} coast \citep{wustrowerosion2025}, about 0.6\,km from Wustrow and counted as the same site, enters only the indicator set of Section~\ref{sec:proxies}. Sellin and Wustrow, together with the Turnpike and
Malaysia sample surveys of Wingtra \citep{wingtra}, extend the re-mapped hold-out of
Section~\ref{sec:remap}.

Seven of the ten captures have RTK-fixed positions, and the two Wingtra surveys have the
positions supplied with the sample data. For Helenenschacht and both Tuniu
captures, the image metadata report a fixed solution for every frame, with standard
deviations of 1.0 to 1.4\,cm horizontally and 2.4 to 3.0\,cm vertically. Bellus has plain
geotags without RTK information, so its reference is accurate to metres only. Its sparsity
results lie within that uncertainty, so its role is mainly in the match-corruption experiment,
where its errors are far larger.

The public data have independent reference poses. ETH3D \citep{eth3d} provides 13 scenes
with laser-scanner references, and EuRoC MH\_01 \citep{burri2016euroc} a drone flight with
total-station positions. The 30 labelled training scenes of IMC 2025 \citep{imc2025} have
benchmark poses. DTU \citep{dtu} is excluded, because no withheld image on its wide-baseline rings could be re-localised.

\begin{table}[t]
\caption{Data, reference and the result sections that use each set.}
\label{tab:data}
\centering
\small
\setlength{\tabcolsep}{3pt}
\begin{tabular}{@{}lrll@{}}
\toprule
Data & Images & Reference & Used in \\
\midrule
Helenenschacht & 176 & RTK, 5 checks & 5.1--5.6 \\
Tuniu 0916 & 297 & RTK & 5.1--5.6 \\
Tuniu 0411 & 271 & RTK & 5.1--5.6 \\
Bellus & 122 & GNSS (metres) & 5.1--5.6 \\
GUT Campus & 612 & RTK & 5.2, 5.5 \\
Sellin & 722 & RTK & 5.1, 5.4, 5.6 \\
Wustrow & 349 & RTK & 5.4, 5.6 \\
Niehagen & 555 & RTK & 5.6 \\
Turnpike & 649 & GNSS & 5.4 \\
Malaysia & 779 & GNSS & 5.4 \\
ETH3D, 13 scenes & & laser scanner & 5.2 \\
EuRoC MH\_01 & 80 & total station & 5.2 \\
IMC 2025, 30 scenes & 9--200 & benchmark & 5.7 \\
\bottomrule
\end{tabular}
\end{table}

\subsection{Accuracy reference}\label{sec:reference}
All reconstructions were computed with COLMAP \citep{colmap} without using GNSS. To
measure their accuracy, we aligned the camera centres to the GNSS positions with a
seven-parameter similarity transformation \citep{umeyama} and report the RMSE of the
remaining differences. The alignment removes scale, rotation and translation, so this
RMSE measures the shape error of the camera block, together with the noise of the
reference. It does not test absolute georeferencing or a uniform scale error.
This least-squares fit has no protection against outliers. A few cameras placed far from the block can shrink its scale, and the RMSE then
approaches the size of the block whatever the rest of the model does. In one Sellin run, three cameras reduce the scale to nearly zero and give 176\,m
RMSE, although 534 of 602 cameras lie within 5\,m under a robust fit. A large least-squares RMSE still indicates lost cameras, but it does not measure block deformation. For corrupted models we therefore also report a robust evaluation: a least-median-of-squares
similarity over 3,000 random triples, refined on the best 80\,\% of the cameras, with the RMSE$_{80}$ of those cameras and the number of cameras within 5\,m. We call a
reconstruction nominal when it comes from the unmodified data, not when its error is
below an application tolerance.

At Helenenschacht we compared camera-position error with the accuracy of object
points. The site has five surveyed ground control points with 35 manual image
measurements. No reconstruction uses them, so they serve as check points. We triangulated
them in eleven reconstructions, nominal and corrupted, and compared them with the survey
after a robust similarity fitted to the cameras. The three runs of the current
database reach 0.08\,m horizontal RMSE. The marker measurements have
reprojection residuals of about 5\,pixels, or 6\,cm at the ground sampling distance of
12.1\,mm, so this value is of the same order as the marking uncertainty. The check points also show a
mean vertical offset, 0.23\,m in these three and 1.5 to 1.9\,m in the eight models of an earlier database.
It follows the focal length of the model (Pearson $r=-0.99$ over the eleven models, focal lengths of 4068 to 4214\,pixels),
at $-0.011$\,m per pixel, close to the $-0.012$\,m expected from the flying height over the focal length.
This is the known coupling of focal length and depth in a nadir block. After removing the mean offset, the 3D RMSE is 0.08 to 0.13\,m in all eleven.
The corrupted model of Table~\ref{tab:failure} at $f=0.8$ shows why the alignment matters. With the least-squares fit its
camera RMSE is 13.5\,m and its check-point RMSE 10.9\,m, and the two errors have a Spearman correlation of 1.00 over the eleven models.
With the robust fit its check-point RMSE is 1.53\,m, the same as the other models of its database. The check points lie in the intact part
of the block, and the shared least-squares fit produced the agreement. Only one site has check points, and none of its models is wrong in the part that they cover.

\subsection{Degradation experiments}\label{sec:degradation}
To vary reconstruction quality, we re-ran the COLMAP mapper from a stored database of
features and matches, which takes seconds to minutes per run. We applied two degradations. In the sparsity experiment, only every $k$-th image in capture
order is kept. In the match-corruption experiment, a fraction $f$ of the verified image
pairs with at least eight matches is chosen at random, and the keypoint indices of one
image in each chosen pair are permuted. The stored two-view geometry of the pair is kept,
so the mapper still treats the wrong correspondences as verified. A third degradation,
which folds part of a survey onto another part with false matches, is described in
Section~\ref{sec:proxies}.

\section{Results}\label{sec:results}

\subsection{Nominal reconstructions and sensitivity}\label{sec:correct}\label{sec:sens}
The nominal reconstructions in this section are the models delivered by the processing
pipeline. The experiments of Sections~\ref{sec:sparsity} to \ref{sec:remap} and \ref{sec:proxies} run the mapper again from stored databases of two generations.
Sections~\ref{sec:sparsity} and \ref{sec:remap} use the current ones, with about 12,500
keypoints per image and camera models from simple radial to full OpenCV. Table~\ref{tab:failure} and
Section~\ref{sec:proxies} use an earlier generation, which on the Tuniu captures has about 20,000 keypoints
per image, 2.5 times as many verified pairs and a simple radial camera. Across the two generations, the nominal
camera-position RMSE is 0.53 and 0.08\,m at Helenenschacht, 1.39 and 1.21\,m at Tuniu 0916,
1.44 and 0.11\,m at Tuniu 0411, and 3.43 and 1.35\,m at Bellus. We did not isolate which change
causes this. The standard score is 1.00 on all eight of these re-run models, despite a 13-fold difference in error at Tuniu 0411. On the delivered models, the angular errors are very small.
At Helenenschacht the median $e^R$ is 0.0025\dg\ and the median $e^t$ is 0.0005\dg.

To check for false alarms on nominal models, we repeated the evaluation
with ten further hold-out seeds on each of the four core captures, 40 runs in total. The
score was 1.00 in every run, with per-seed median rotation errors of 0.003\dg\ to
0.027\dg. These splits of the same models are not independent trials and give no false-alarm rate. The delivered Bellus model, a separate run, scores 0.97 with the default seed because one of its ten withheld images misses the
1\dg\ threshold.

We also tested how the score depends on the design choices of the protocol (tables in Online Resource 1). Between hold-out fractions of 0.05
and 0.5, the score stays between 0.93 and 1.00 on the four core captures and Sellin. Raising the threshold from $\tau=2$ to $\tau=5$
increases the median rotation error by a factor of 5 to 15, to at most 0.11\dg, while the score stays at 0.89 or above. At $\tau=5$ the coverage falls to 0.705 at Sellin.

The distance between the resected and the stored projection centre needs no alignment and does not depend on the origin. Its median is
at most $2.7\times10^{-4}$ of the RMS extent of the camera block on the five captures, so a
variant based on it is as saturated.

\subsection{Saturation under image sparsity}\label{sec:sparsity}
\begin{figure}[t]
\centering
\includegraphics[width=\columnwidth]{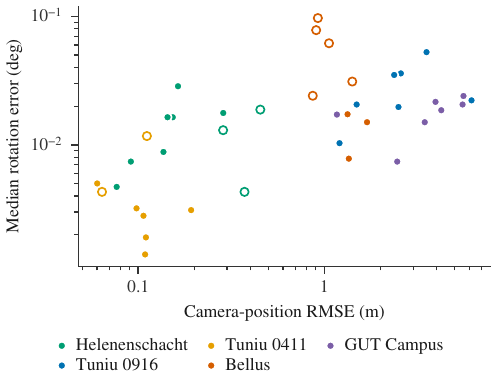}
\caption{Median rotation error of the withheld images against
camera-position RMSE for every level of the sparsity experiment, all with a score of 1.00. Open circles mark
sub-models with fewer than ten registered images.}
\label{fig:sparsity}
\end{figure}

We thinned the four core captures and GUT Campus to every $k$-th image, with $k$ from 1
to 20, and kept the 40 levels that gave a score (Fig.~\ref{fig:sparsity},
per-capture values in Online Resource 1). The core captures were thinned in one run and GUT Campus in a separate one. The score was 1.00 in all 40 levels, while the camera-position RMSE
varied by a factor of up to 5.9 within one capture. It ranged from 0.077 to 0.453\,m at Helenenschacht, 1.21 to 6.17\,m at Tuniu 0916, 0.06 to 0.19\,m at Tuniu 0411, 0.87 to 1.70\,m at Bellus and 1.17 to 5.59\,m at GUT Campus.

The largest errors among models of at least ten images, 6.2\,m at Tuniu 0916 and 5.6\,m at
GUT Campus, also scored 1.00. Scores below 1.00 do occur, but they did not track the error or recur in repeated runs. An
earlier sweep on the older databases gave three in 43 levels. A four-image sub-model scored 0.67, a Tuniu 0916 model of 43 images scored 0.92
with about 20\,m RMSE, and a Tuniu 0411 model of 56 images scored 0.83 with about 4\,m. In a repeat of the Tuniu 0916 sweep, the same 43-image level had 2.3\,m RMSE and scored 1.00, while a model of 85 images with 1.5\,m RMSE, one of the most accurate of the sweep, scored 0.96.

Two reasons explain the saturation. Thresholds of 1\dg\ and more are far coarser
than the observed median errors, which are all below 0.1\dg, and thinning makes a model smaller but not inconsistent. The RMSE values also need care: models of 3 to 5 images rest on nearly exact similarity fits, and Bellus has only a metre-level reference.

The score also saturates against independent ground truth. On 106 degraded reconstructions of the
13 ETH3D scenes, 56 thinned and 50 with corrupted matches, the score was 1.00 in every
case, with camera-position errors from 1.5\,mm to 6.6\,m, many of the largest in small sub-models. Among larger models, the facade scene, with 72 images, has 3.4\,m error, and the electro scene, with
40 images, has 1.7\,m and 13.1\dg\ against the reference poses. On EuRoC MH\_01 the score remained at 1.00 in 8 of 10 degraded reconstructions, including one that uses
all 80 images and has 4.3\,m error.

\subsection{Fragmentation, lost cameras and wrong models}\label{sec:failure}
We corrupted the matches of each core capture at $f = 0$, 0.2, 0.5 and 0.8 in one run with
a fixed seed (Table~\ref{tab:failure}). For each model we recorded
the number of models, the registered images, the coverage, the score, and the least-squares and robust errors. Up to $f=0.5$, the outlier rejection of the
mapper absorbs most of the corruption. At $f=0.8$ every capture fails, in three different
ways.

\begin{table*}[t]
\caption{Match-corruption experiment on the four core captures, without the rows for
$f=0.2$. Models is the number of
separate models, Reg.\ the number of images in the scored model, Cov.\ the resected out of
the withheld images that are registered in the scored model, RMSE the least-squares value, RMSE$_{80}$ the robust value of Section~\ref{sec:reference}, and $\le$5\,m the
number of cameras within 5\,m under the robust fit.}
\label{tab:failure}
\centering
\small
\begin{tabular}{@{}lcccccrrrl@{}}
\toprule
Capture & $f$ & Models & Reg. & Cov. & Score & RMSE (m) & RMSE$_{80}$ (m) & $\le$5\,m & Outcome \\
\midrule
Helenenschacht & 0.0 & 1 & 176 & 18/18 & 1.00 & 0.53 & 0.42 & 176 & nominal \\
 & 0.5 & 2 & 175 & 18/18 & 1.00 & 0.66 & 0.53 & 175 & absorbed \\
 & 0.8 & 3 & 169 & 15/17 & 1.00 & 13.47 & 0.46 & 161 & cameras lost \\
Tuniu 0916 & 0.0 & 2 & 294 & 29/29 & 1.00 & 1.39 & 0.86 & 292 & nominal \\
 & 0.5 & 1 & 285 & 27/28 & 1.00 & 12.41 & 0.94 & 272 & cameras lost \\
 & 0.8 & 1 & 257 & 18/26 & 0.96 & 34.30 & 2.70 & 193 & many lost, core degraded \\
Tuniu 0411 & 0.0 & 1 & 271 & 27/27 & 1.00 & 1.44 & 0.89 & 270 & nominal \\
 & 0.5 & 2 & 249 & 25/25 & 1.00 & 3.34 & 0.74 & 246 & absorbed \\
 & 0.8 & 1 & 242 & 20/24 & 0.98 & 41.59 & 1.67 & 190 & many lost, core degraded \\
Bellus & 0.0 & 1 & 105 & 10/10 & 1.00 & 3.43 & 2.25 & 90 & nominal \\
 & 0.5 & 1 & 99 & 10/10 & 0.97 & 4.60 & 2.84 & 75 & absorbed \\
 & 0.8 & 3 & 41 & 2/4 & 1.00 & 88.20 & 46.88 & 0 & wrong throughout \\
\bottomrule
\end{tabular}
\end{table*}

The robust evaluation separates three kinds of failure. In the first, a few cameras are lost and the
core stays intact. Helenenschacht at $f=0.8$ splits into three models, and the largest has 13.5\,m RMSE, but 161 of its 169 cameras lie within 5\,m and its
RMSE$_{80}$ of 0.46\,m is close to the nominal 0.42\,m. Tuniu 0916 at $f=0.5$ behaves in the same way, with 12.4\,m RMSE and 272 of 285 cameras within 5\,m. Both score 1.00. The score agrees with the intact core but gives no indication of the lost cameras.

In the second kind, about a quarter of the cameras is lost and the core degrades. Tuniu 0916 and Tuniu 0411 stay in one
connected model at $f=0.8$, with 34.3 and 41.6\,m RMSE. Under the robust fit 193 of 257 and 190 of 242 cameras lie within 5\,m, and the RMSE$_{80}$ is
2.70 and 1.67\,m, 3.1 and 1.9 times nominal. The
scores are 0.96 and 0.98, and the withheld images that could be resected agree with the
model to a median of 0.05\dg\ and 0.04\dg. In the third kind, the scored part is
wrong throughout. Bellus keeps only 41 images in its largest model, against 105 at $f=0$. None of them lies within 5\,m under the robust fit, and the score is 1.00.
At $f=0.9$, Tuniu 0411 has one of 226 cameras within 5\,m, an RMSE$_{80}$ of 101\,m and a score of 1.00.

The score therefore did not respond to any of the three, but the coverage changed. At $f=0.8$, 2 to 8 withheld images per capture had too few trusted points for a
resection, and counting them as failures (Section~\ref{sec:protocol}) gives 0.88, 0.67,
0.82 and 0.50 for the four captures. A reduced coverage indicates damaged tracks. It stays nearly complete, at 27 of 28, at Tuniu 0916 with $f=0.5$, and
it falls on correct models when the threshold is raised to $\tau=5$
(Section~\ref{sec:sens}).

Separate runs at $f=0.6$, 0.7 and 0.9 confirm the pattern. At $f=0.6$ and 0.7, Tuniu 0411 scores 1.00 with least-squares RMSE of 12.3 and 15.6\,m and RMSE$_{80}$ of 0.79 and 0.87\,m. At $f=0.9$, Tuniu 0916 drops to 0.50, on 2 of 23 withheld images, in a model wrong throughout. At $f=0.9$ Helenenschacht splits
into 13 models, and the largest, with 18 images, gives no score at all. A missing score
and a low coverage must therefore be treated as failures.

\subsection{Re-mapped and re-matched hold-outs}\label{sec:remap}
The trusted points of the standard protocol were adjusted together with the withheld
images. To remove this dependence, we deleted every match that involves an image of
$\mathcal{H}$ and mapped the block again. The withheld images then took no part in
triangulation, bundle adjustment or the refinement of the intrinsics. Each of them was
resected against this retained-only model through its own matches, with at least six correspondences and twelve inliers, stricter minima than in the standard test. A similarity fitted to the shared camera centres
aligned the retained-only model to the full one, and we compared the resected rotation
with the rotation in the full model. This comparison does not depend on the origin or the
scale, but it does depend on the alignment. We call the mAA of this rotation error over 1\dg, 3\dg\ and 5\dg, with failed
resections counted as errors, the re-mapped score. It requires a second mapping run as long as the first, whereas the standard test takes seconds.

We ran it on the nominal database of each core capture and at $f=0.8$ with three
corruption seeds (Table~\ref{tab:remap}). The four nominal models score 0.94
to 1.00. One of the twelve corrupted
runs produced a nearly complete and accurate model, with 0.42\,m RMSE at Helenenschacht, and it scores 1.00.
The other eleven have least-squares errors of 4.0 to 126\,m and score 0.00 to 0.67, while
the standard score is 1.00 in eight of them and 0.75 to 0.92 in the other three.

\begin{table*}[t]
\caption{Hold-out variants on the current databases. Registered images, RMSE and RMSE$_{80}$ in metres, cameras within 5\,m,
standard score $s$ and coverage of the full model, re-matched score $s_m$, re-mapped score $s_r$ and control score $s_c$ with
least-squares alignment, and $s_r^\ast$ and $s_c^\ast$ with robust alignment.}
\label{tab:remap}
\centering
\small
\begin{tabular}{@{}lcrrrrccccccc@{}}
\toprule
 & & \multicolumn{6}{c}{Full model} & & \multicolumn{2}{c}{Re-mapped} & \multicolumn{2}{c}{Control} \\
\cmidrule(lr){3-8}\cmidrule(lr){10-11}\cmidrule(l){12-13}
Capture & $f$ & Reg. & RMSE & RMSE$_{80}$ & $\le$5\,m & $s$ & Cov. & $s_m$ & $s_r$ & $s_r^\ast$ & $s_c$ & $s_c^\ast$ \\
\midrule
Helenenschacht & 0.0 & 176 & 0.08 & 0.06 & 176 & 1.00 & 18/18 & 1.00 & 1.00 & 1.00 & 1.00 & 1.00 \\
 & 0.8 & 169 & 0.42 & 0.09 & 169 & 1.00 & 17/17 & 1.00 & 1.00 & 1.00 & 0.67 & 1.00 \\
 & 0.8 & 169 & 8.24 & 0.11 & 165 & 1.00 & 17/17 & 1.00 & 0.35 & 0.59 & 0.53 & 0.69 \\
 & 0.8 & 167 & 4.81 & 0.10 & 165 & 1.00 & 17/17 & 0.94 & 0.67 & 1.00 & 0.98 & 0.98 \\
Tuniu 0916 & 0.0 & 294 & 1.21 & 0.87 & 294 & 1.00 & 29/29 & 1.00 & 0.94 & 0.94 & 1.00 & 1.00 \\
 & 0.8 & 248 & 62.35 & 1.97 & 193 & 1.00 & 17/25 & 0.65 & 0.00 & 0.56 & 0.28 & 0.76 \\
 & 0.8 & 261 & 50.88 & 3.56 & 186 & 0.92 & 25/26 & 0.86 & 0.01 & 0.60 & 0.00 & 0.76 \\
 & 0.8 & 243 & 125.67 & 9.67 & 127 & 0.90 & 17/24 & 0.56 & 0.00 & 0.36 & 0.00 & 0.40 \\
Tuniu 0411 & 0.0 & 271 & 0.11 & 0.08 & 271 & 1.00 & 27/27 & 1.00 & 1.00 & 1.00 & 1.00 & 1.00 \\
 & 0.8 & 246 & 23.54 & 0.08 & 231 & 1.00 & 25/25 & 1.00 & 0.60 & 0.92 & 0.00 & 1.00 \\
 & 0.8 & 14 & 4.03 & 2.67 & 12 & 1.00 & 1/1 & 1.00 & 0.67 & 1.00 & 0.00 & 0.00 \\
 & 0.8 & 156 & 8.34 & 0.07 & 154 & 1.00 & 16/16 & 0.88 & 0.38 & 0.38 & 1.00 & 1.00 \\
Bellus & 0.0 & 98 & 1.35 & 1.04 & 98 & 1.00 & 10/10 & 1.00 & 1.00 & 1.00 & 1.00 & 1.00 \\
 & 0.8 & 46 & 28.62 & 2.42 & 35 & 1.00 & 4/5 & 0.80 & 0.00 & 0.40 & 0.00 & 0.73 \\
 & 0.8 & 36 & 37.28 & 4.12 & 26 & 1.00 & 3/4 & 0.67 & 0.00 & 0.00 & 0.40 & 0.60 \\
 & 0.8 & 35 & 22.37 & 2.36 & 28 & 0.75 & 4/4 & 0.75 & 0.00 & 0.00 & 0.00 & 0.58 \\
\bottomrule
\end{tabular}
\end{table*}

Much of this response comes from the alignment. In several runs all withheld images appeared to be off by the same rotation, 2.4\dg\ to 2.9\dg\ at Helenenschacht and about 64\dg\ at Tuniu 0916,
and the retained cameras show the same offset under the same fit. The withheld images agree with the retained-only model. Lost cameras in either model tilt the
least-squares alignment between the two, and the score then measures that tilt. A low score therefore indicates that one of the models has lost cameras, but it is not a measured rotation error of the withheld images. When we score the
same resected poses after a robust alignment ($s_r^\ast$), one Helenenschacht run rises from 0.67 to 1.00 and one Tuniu 0411 run from 0.60 to 0.92. Of the 17 corrupted runs with a least-squares error above 4\,m on the seven captures, five then score above 0.67. The rest stay lower, because of fragmentation or a real difference in shape.

To separate the effect of withholding from the behaviour of the mapper, we repeated every
case without withholding anything. The full database was mapped twice, and the poses of
the same images in the two models were compared in the same way ($s_c$ in
Table~\ref{tab:remap}). On the nominal databases the two runs agree to a median of
at most 0.03\dg\ and score 1.00. On most corrupted databases the two runs disagree under either alignment, which shows that these databases do not determine one solution. A second mapping run exposes this, and withholding the images adds little.

We repeated the experiment on Wustrow, Sellin and Turnpike, and on the nominal database of
Malaysia. For these extra sites, before computing correlations, we required both models of a run to register at least half
of the images, which left 10 hold-out and 8 control runs. Three further captures gave no usable
result. The default mapper registered 2 of 191 images at Luverne and kept 19 of 680 at a bridge survey, and at EuroSDR UK no pair of models both reached half of the 430 images. The nominal models
at Malaysia, Turnpike and Sellin have 0.2 to 1.6\,m RMSE and re-mapped scores of
0.99 to 1.00. Nominal Wustrow has 9.2\,m RMSE, 6.0\,m RMSE$_{80}$, a standard score of 1.00 and $s_r=0.71$. We did not establish the cause of its error. The
six corrupted runs have 21 to 176\,m least-squares RMSE, but the two at Sellin and the one at Turnpike keep RMSE$_{80}$ near nominal. Their standard scores are 0.87 to 1.00. At Turnpike the control scores 0.94 and 0.95 although
its two runs have 21 and 38\,m RMSE, so agreeing rotations do not exclude lost cameras.

Taking the correspondences from the verified matches gives a third test. For each keypoint of a withheld image, its verified matches to retained images select a 3D point by voting. Ties are dropped, and the image's own assignments are not used. The track-length threshold, the resection and the score are as in the standard test. It takes seconds on a saved model and the database, and needs no second mapping run. We computed this re-matched score $s_m$ on 25 saved runs. All seven nominal models score 1.00, with median errors of 0.002\dg\ to 0.02\dg. Of the 18 corrupted runs,
14 fall below 0.95, down to 0.54, against 4 of 18 for the standard score on the same models and 10 of 18 when its failed resections count as errors ($c\,s$). Using the uncorrupted match table gives a similar range of 0.53 to 1.00, so the drop reflects the model and not only the scrambled evidence. It does not flag nominal Wustrow, which has 9.2\,m RMSE. Three corrupted runs with least-squares errors of 4 to 24\,m also score 1.00, two of them with an intact core. This result rests on one kind of corruption and needs testing on natural failures.

Table~\ref{tab:scoreboard} compares the scores against three error measures. Runs of one capture are not independent, so these values only describe the sample.
The least-squares RMSE favours $s_r$, partly because the same outliers disturb both. Against RMSE$_{80}$ and the fraction of lost cameras, the differences among $c\,s$, $s_m$, $s_r$ and $s_r^\ast$ are small, $s$ is lowest, and the control with robust alignment is highest or tied. Most of the gain over $s$ therefore comes from counting failed resections, and the re-mapped test has no clear advantage over the cheaper alternatives. A bootstrap over the seven captures gives wide 95\,\% intervals, for example $-0.88$ to $-0.57$ for $c\,s$ against RMSE$_{80}$, so these differences are not resolved. Comparing two mapping runs is informative mainly because it tests reproducibility.

\begin{table}[t]
\caption{Spearman correlation of each score with error over 25 hold-out runs on seven captures and 23 control runs, without Malaysia, whose models were not kept. Lost is the fraction of cameras farther than 5\,m under the robust fit.}
\label{tab:scoreboard}
\centering
\small
\begin{tabular}{@{}lccc@{}}
\toprule
Score & RMSE & RMSE$_{80}$ & Lost \\
\midrule
Standard $s$ & $-0.69$ & $-0.60$ & $-0.60$ \\
Standard $c\,s$ & $-0.81$ & $-0.75$ & $-0.78$ \\
Re-matched $s_m$ & $-0.80$ & $-0.67$ & $-0.75$ \\
Re-mapped $s_r$ & $-0.92$ & $-0.66$ & $-0.78$ \\
Re-mapped $s_r^\ast$ & $-0.78$ & $-0.63$ & $-0.74$ \\
Control $s_c$ & $-0.89$ & $-0.68$ & $-0.87$ \\
Control $s_c^\ast$ & $-0.79$ & $-0.83$ & $-0.87$ \\
\bottomrule
\end{tabular}
\end{table}

\needspace{8\baselineskip}
\subsection{Finer thresholds}\label{sec:desat}
Part of the saturation comes from thresholds that are much coarser than the errors. The
median rotation error of the withheld images, used without a threshold
(Fig.~\ref{fig:sparsity}), correlates with RMSE at 0.20 (95\,\% CI $-0.12$ to 0.48)
across the 40 sparsity reconstructions, and at 0.53 across the 30 with at least ten images. These intervals treat nested levels as independent and are too narrow, and a random-effects meta-analysis over the five captures gives 0.02. A standard score with
thresholds of 0.03\dg, 0.1\dg\ and 0.3\dg\ falls from 0.85--0.91 on the nominal RTK
models of Table~\ref{tab:failure} to 0.63 and 0.62 on Tuniu 0916 and Tuniu 0411 at $f=0.8$. However, the nominal Bellus model scores 0.67. The mean rotation errors of these two models, 0.22\dg\ and 0.13\dg\ (medians 0.05\dg\ and 0.04\dg{}), are close to the medians of up to 0.11\dg\ on nominal models under the stricter threshold $\tau=5$, so none of the sub-degree thresholds we tested separated the two groups.

\subsection{Other internal indicators}\label{sec:proxies}
If the errors are absorbed locally, other internal indicators should miss them as well.
We compared four of them on the corrupted models: reprojection RMSE,
the standard hold-out score $s$ without the coverage, the number of registered
images and the mean track length. For each, the area under the ROC curve (AUROC) measures how well it
separates catastrophic from acceptable models. A value of 0.5 means chance, and a value
below 0.5 means that the indicator ranks the catastrophic models as better. Each indicator is
oriented so that the value expected of a worse model is higher, for example a larger
reprojection error or a shorter track. Rank correlations with error use the same
orientation.

\begin{table}[t]
\caption{AUROC of internal indicators for separating catastrophic from acceptable
corrupted models. Set A is the 16 models of the experiment of Table~\ref{tab:failure} (catastrophic above
5\,m least-squares RMSE), set B 19 models from a separate corruption run on Sellin, Wustrow and Niehagen, each processed twice (above 20\,m).}
\label{tab:proxies}
\centering
\small
\begin{tabular}{@{}lcc@{}}
\toprule
Indicator & Set A & Set B \\
\midrule
Reprojection RMSE & 0.15 & 0.10 \\
Hold-out score & 0.65 & 0.91 \\
Registered images & 0.55 & 0.64 \\
Mean track length$^{a}$ & 0.89 & 0.97 \\
\bottomrule
\end{tabular}

\smallskip
\raggedright\footnotesize
Set A has 5 catastrophic models and set B has 9.
$^{a}$The AUROC is inflated, because corruption shortens tracks and increases error at the same time.
\end{table}

On set A (Table~\ref{tab:proxies}), reprojection RMSE has an AUROC of
0.15 and a Spearman correlation with error of $-0.40$.
The catastrophic models have lower reprojection errors than the nominal ones, so the most
widely reported quality figure ranks some of the worst models as best. The hold-out score reaches 0.65 on set A and 0.91 on set B, the
best of the consistency-based indicators there. The inversion of reprojection RMSE does not depend on the least-squares label. Labelling models by RMSE$_{80}$ or by more than 10\,\% lost cameras gives 0.07 and 0.32 on set A and 0.15 and 0.18 on set B, while the hold-out score ranges from 0.40 to 0.75 on set A and from 0.71 to 0.81 on set B. The samples are small, and set B comes from only three surveys.

Mean track length reaches 0.89 and 0.97, but these values are inflated by the design of
the experiment. To separate damage to the
tracks from a displaced part of the block, we used a second kind of corruption. False matches
fold a cluster of images onto another part of the survey. All conditions used the same number of folded images, false matches per pair and anchors. Only the position of the fold varied. At Bellus, the least-squares RMSE across these conditions ranged from 3.5 to 85.1\,m,
while the mean track length changed by 0.5\,\%. The RMSE$_{80}$ stayed at 2.3\,m: the fold displaces the folded cluster and leaves the rest of the block in place. The fold took effect at three of five sites:
Bellus, Wustrow and Tuniu 0411. There, the within-site rank correlations of track length
with error had no consistent sign ($-0.80$, 0.80 and $-0.40$). Reprojection RMSE behaved similarly, and the hold-out score was constant at
two of the three sites. Track length thus responds to damaged tracks but not to a
displacement that leaves the tracks intact.

\subsection{Benchmark scenes}\label{sec:imc}
To test the score on independent scenes of naturally varying difficulty, we reconstructed
all 30 labelled training scenes of IMC 2025 with a standard COLMAP pipeline. We compared
the hold-out score with a camera-centre mAA against the released poses. We reimplemented the pose term of the official scoring: centres aligned by a similarity found by RANSAC over triplets of
cameras, an image correct at each of the six thresholds of its scene when its
centre lies within it, and unregistered images counted as errors.

Seven scenes failed to reconstruct, an eighth registered only 2 of 85 images, too few to withhold any, and a ninth had no withheld image that could be evaluated. With a missing score counted as zero, score and
benchmark mAA have a Spearman correlation of 0.68 (bootstrap 95\,\% CI 0.33 to 0.87). This
comes almost entirely from separating failed from successful scenes. Among the 21
scenes with an evaluable score, the score lies between 0.95 and 1.00 and has no clear rank correlation
with mAA ($\rho=-0.23$, CI $-0.58$ to 0.14), although mAA ranges from 0.11 to 0.75.  Nine of the 21 scenes have fewer
than three withheld images that can be evaluated. On the 12 larger scenes, $\rho$ is
$-0.20$.

We also reconstructed the 16 scenes of six IMC datasets with five pipeline configurations. The mean benchmark mAA per setting
ranged from 0.21 to 0.53. Of the 80 reconstructions, 72 gave a score, and 60 of these scored 1.00. Ordering pairs of
configurations by benchmark mAA, the score gave the correct order in 51\,\% of cases, the
level of chance, against 66\,\% for the median rotation error and
76\,\% for the number of registered images.  The
score must be read together with the registration rate.

\section{Discussion}\label{sec:discussion}
In terms of reliability theory, the standard hold-out score measures the internal
consistency of one reconstruction. It did not respond to errors that the block absorbs,
and these are the errors that matter for external reliability. Models with a quarter of their cameras lost scored
0.96 to 0.98, a model wrong throughout scored 1.00, and reprojection error ranked some of the worst
models best. Finer thresholds did not separate these from correct models of weaker blocks.

Part of the saturation is due to how the test is built. Counting failed resections as errors lowered the score below 0.95 on 10 of 18 corrupted runs, and reading correspondences from the verified matches on 14. Mapping again also responded, and the control shows why: the corrupted databases did
not determine one solution. Two mapping runs of the same data, with or without the
withheld images, gave different models, while the nominal databases gave nearly the same
model. Comparing two runs requires one more mapping run and no ground truth. Any comparison of two models must use a robust alignment, or a few lost cameras will dominate it. Unlike \citet{lookma2024}, it needs no explicit perturbation of the input, because mapping is stochastic. It says nothing about an error that every run repeats, such as a wrong camera
model, doming, scale drift along a sequence or a wrong merge of repeated structures
\citep{heinly2015world}. A model that reconstructs the same way twice can still be wrong. The focal-length offset at Helenenschacht is an example: three runs of one database agree to millimetres and keep a 0.23\,m vertical offset at the check points. Across all eleven models the offset ranges from 0.23 to 1.9\,m, and the standard score is 1.00 on every one of them.

For quality control of inspection products, three consequences follow. First, none of the
internal indicators we tested can replace check points in an accuracy statement.
Positional accuracy standards base such statements on independent check points
\citep{asprs2023}, and our results support this requirement. Second, the
score with its coverage and the registration rate is a cheap screen for fragmentation. Failed resections should count as errors, correspondences from matches add some sensitivity, and a second mapping run with robust alignment is a screen for unstable solutions. The alarm levels of these screens are not calibrated, and the scores overlap between accurate and failed runs. A failed screen therefore calls for inspection, not for automatic rejection. Reports of alignment RMSE should state how many cameras dominate it. Third,
the errors that pass every internal check can be found only with information from outside
the block. This can be withheld GNSS camera positions, a scale bar, or
prior knowledge such as vertical walls. Where full ground control is too costly, a few check points provide this information at low cost \citep{gerke2016accuracy}.

\section{Limitations}\label{sec:limitations}
The number of independent sites is small, and correlations at this sample size can detect only strong effects.

Both degradation experiments vary one parameter that drives the degradation and the error
together, which can create correlations between indicators and error. The IMC experiments contain no injected degradation.

The re-mapped and re-matched scores rest on one corruption
level and on seven or eight captures, on databases that differ from those of the other experiments. The robust evaluation has its own limits: it assumes that the best 80\,\% of the cameras are a correct core, and it is less meaningful when a model is wrong throughout. The thinning models were not kept, so the sparsity experiment has least-squares RMSE only. We did not turn these scores into a detector with a chosen threshold and measured error
rates, which would need many more sites and natural failures.

Point accuracy was checked at one site only, and none of its eleven models is wrong in the part that the check points cover. Public benchmarks such as UseGeo \citep{usegeo2024} would allow this check to be repeated. The failed models come from injected
corruption. A naturally occurring merge of repeated structures
\citep{heinly2015world} with survey ground truth is not included. How often such
failures occur in routine work remains unknown.

\section{Conclusion}\label{sec:conclusion}
We studied hold-out self-validation of photogrammetric reconstructions without ground
truth. The standard protocol re-localises each withheld image against points observed in other images, using the correspondences assigned by the mapper. Its score measures internal consistency and saturates. A model with one of 226 cameras within 5\,m
scored 1.00, and reprojection error did no better.
Counting failed resections as errors and reading correspondences from the verified matches both respond on most corrupted models. Mapping the block again also responds,
mainly because corrupted reconstructions do not reproduce from run to run, and it needs a robust alignment. Least-squares alignment RMSE can overstate the error of a corrupted model by two orders of magnitude and should not be read as deformation. An error that
every reconstruction repeats passes all these tests. Check points or another independent
measurement remain necessary.

\section*{Statements and Declarations}

\subsection*{Competing interests}
\ifanon
The author has filed a patent application relating to the protocol described here
(details withheld for anonymous review).
\else
The author has filed a U.S. provisional patent application (No.~64/113,075) relating to
the protocol described here.
\fi There are no other competing financial or non-financial
interests to declare.

\subsection*{Funding}
The author received no funding, institutional support or third-party support for this
work.

\subsection*{Data and code availability}
The hold-out evaluator, the degradation and analysis scripts, the worked example and the
per-reconstruction results behind every figure and table are archived at Zenodo under the
MIT licence (\href{https://doi.org/10.5281/zenodo.21737747}{10.5281/zenodo.21737747}). The earlier reconstructions were computed with pycolmap 4.0. The re-mapped, re-matched and thinning experiments and the re-evaluation of the archived models, with PROJ-based geodesy and robust alignment, used pycolmap 4.2, which reproduces the
archived scores. ETH3D, EuRoC and IMC 2025 are available
from their providers. All ten aerial captures are public (Section~\ref{sec:datasets}). The archive contains the
score, RMSE, coverage and registered images of each reconstruction, which suffice to
reproduce every reported statistic. The archive also contains the earlier runs mentioned in the text.
\ifanon\else
An earlier version of this work is available as arXiv preprint
\href{https://arxiv.org/abs/2607.24852}{2607.24852}.
\fi

\subsection*{Author contributions}
\ifanon The author\else Behnam Asadi\fi{} conceived the study, designed and implemented the
protocol and the degradation experiments, carried out all experiments and analyses, prepared
all figures and tables, and wrote the manuscript.

\subsection*{Use of large language models}
The author used AI tools for language editing and polishing. The author verified all results and takes full responsibility for the content.

\bibliographystyle{spbasic}
\bibliography{refs}

\begin{thebibliography}{37}
\providecommand{\natexlab}[1]{#1}
\providecommand{\url}[1]{{#1}}
\providecommand{\urlprefix}{URL }
\expandafter\ifx\csname urlstyle\endcsname\relax
  \providecommand{\doi}[1]{DOI~\discretionary{}{}{}#1}\else
  \providecommand{\doi}{DOI~\discretionary{}{}{}\begingroup
  \urlstyle{rm}\Url}\fi
\providecommand{\eprint}[2][]{\url{#2}}

\bibitem[{Aan{\ae}s et~al.(2016)Aan{\ae}s, Jensen, Vogiatzis, Tola, and
  Dahl}]{dtu}
Aan{\ae}s H, Jensen RR, Vogiatzis G, Tola E, Dahl AB (2016) Large-scale data
  for multiple-view stereopsis. International Journal of Computer Vision
  120(2):153--168, \doi{10.1007/s11263-016-0902-9}

\bibitem[{{American Society for Photogrammetry and Remote
  Sensing}(2023)}]{asprs2023}
{American Society for Photogrammetry and Remote Sensing} (2023) {ASPRS}
  positional accuracy standards for digital geospatial data, edition 2, version
  1.0. ASPRS, Bethesda, MD

\bibitem[{Baarda(1968)}]{baarda1968}
Baarda W (1968) A Testing Procedure for Use in Geodetic Networks, Publications
  on Geodesy, New Series, vol~2. Netherlands Geodetic Commission, Delft

\bibitem[{Burdziakowski(2026)}]{gutcampus}
Burdziakowski P (2026) {UAV} survey images -- {3D} model -- {GUT} full campus
  -- share {102S}. \doi{10.34808/jk93-y864}, dataset

\bibitem[{Burri et~al.(2016)Burri, Nikolic, Gohl, Schneider, Rehder, Omari,
  Achtelik, and Siegwart}]{burri2016euroc}
Burri M, Nikolic J, Gohl P, Schneider T, Rehder J, Omari S, Achtelik MW,
  Siegwart R (2016) The {EuRoC} micro aerial vehicle datasets. The
  International Journal of Robotics Research 35(10):1157--1163,
  \doi{10.1177/0278364915620033}

\bibitem[{Fontan et~al.(2024)Fontan, Civera, Fischer, and Milford}]{lookma2024}
Fontan A, Civera J, Fischer T, Milford M (2024) Look ma, no ground truth!
  {G}round-truth-free tuning of structure from motion and visual {SLAM}. arXiv
  preprint, \url{https://arxiv.org/abs/2412.01116}

\bibitem[{F{\"o}rstner(1987)}]{forstner1987reliability}
F{\"o}rstner W (1987) Reliability analysis of parameter estimation in linear
  models with applications to mensuration problems in computer vision. Computer
  Vision, Graphics, and Image Processing 40(3):273--310

\bibitem[{Frahm and Pollefeys(2006)}]{frahm2006qdegsac}
Frahm JM, Pollefeys M (2006) {RANSAC} for \mbox{(quasi-)degenerate} data
  ({QDEGSAC}). In: IEEE Conf. on Computer Vision and Pattern Recognition (CVPR)

\bibitem[{Gerke and Przybilla(2016)}]{gerke2016accuracy}
Gerke M, Przybilla HJ (2016) Accuracy analysis of photogrammetric {UAV} image
  blocks: influence of onboard {RTK-GNSS} and cross flight patterns.
  Photogrammetrie -- Fernerkundung -- Geoinformation (1):17--30,
  \doi{10.1127/pfg/2016/0284}

\bibitem[{Heinly et~al.(2015)Heinly, Sch{\"o}nberger, Dunn, and
  Frahm}]{heinly2015world}
Heinly J, Sch{\"o}nberger JL, Dunn E, Frahm JM (2015) {Reconstructing the
  World* in Six Days *(As Captured by the Yahoo 100 Million Image Dataset)}.
  In: CVPR

\bibitem[{{Image Matching Challenge organisers}(2025)}]{imc2025}
{Image Matching Challenge organisers} (2025) Image matching challenge 2025.
  Kaggle competition,
  \url{https://www.kaggle.com/competitions/image-matching-challenge-2025},
  accessed 2026-07-15

\bibitem[{James and Robson(2014)}]{james2014mitigating}
James MR, Robson S (2014) Mitigating systematic error in topographic models
  derived from {UAV} and ground-based image networks. Earth Surface Processes
  and Landforms 39(10):1413--1420, \doi{10.1002/esp.3609}

\bibitem[{James et~al.(2017)James, Robson, and Smith}]{james2017precision}
James MR, Robson S, Smith MW (2017) {3-D} uncertainty-based topographic change
  detection with structure-from-motion photogrammetry: precision maps for
  ground control and directly georeferenced surveys. Earth Surface Processes
  and Landforms 42(12):1769--1788, \doi{10.1002/esp.4125}

\bibitem[{Jin et~al.(2021)Jin, Mishkin, Mishchuk, Matas, Fua, Yi, and
  Trulls}]{imc}
Jin Y, Mishkin D, Mishchuk A, Matas J, Fua P, Yi KM, Trulls E (2021) Image
  matching across wide baselines: From paper to practice. Int Journal of
  Computer Vision (IJCV) 129:517--547

\bibitem[{Kendall and Gal(2017)}]{kendall2017uncertainties}
Kendall A, Gal Y (2017) What uncertainties do we need in {Bayesian} deep
  learning for computer vision? In: Advances in Neural Information Processing
  Systems (NeurIPS)

\bibitem[{Knapitsch et~al.(2017)Knapitsch, Park, Zhou, and Koltun}]{tanks}
Knapitsch A, Park J, Zhou QY, Koltun V (2017) Tanks and temples: Benchmarking
  large-scale scene reconstruction. ACM Trans on Graphics 36(4)

\bibitem[{Kneip et~al.(2011)Kneip, Scaramuzza, and Siegwart}]{kneip2011p3p}
Kneip L, Scaramuzza D, Siegwart R (2011) A novel parametrization of the
  perspective-three-point problem for a direct computation of absolute camera
  position and orientation. In: CVPR

\bibitem[{K{\"o}lle et~al.(2021)K{\"o}lle, Laupheimer, Schmohl, Haala,
  Rottensteiner, Wegner, and Ledoux}]{h3d2021}
K{\"o}lle M, Laupheimer D, Schmohl S, Haala N, Rottensteiner F, Wegner JD,
  Ledoux H (2021) The {Hessigheim 3D} ({H3D}) benchmark on semantic
  segmentation of high-resolution {3D} point clouds and textured meshes from
  {UAV} lidar and multi-view-stereo. ISPRS Open Journal of Photogrammetry and
  Remote Sensing 1:100001

\bibitem[{Lepetit et~al.(2009)Lepetit, Moreno-Noguer, and
  Fua}]{lepetit2009epnp}
Lepetit V, Moreno-Noguer F, Fua P (2009) {EPnP}: An accurate {$O(n)$} solution
  to the {PnP} problem. International Journal of Computer Vision 81(2):155--166

\bibitem[{Luhmann et~al.(2016)Luhmann, Fraser, and Maas}]{luhmann2016sensor}
Luhmann T, Fraser C, Maas HG (2016) Sensor modelling and camera calibration for
  close-range photogrammetry. ISPRS Journal of Photogrammetry and Remote
  Sensing 115:37--46, \doi{10.1016/j.isprsjprs.2015.10.006}

\bibitem[{Nex et~al.(2015)Nex, Gerke, Remondino, Przybilla, B{\"a}umker, and
  Zurhorst}]{nex2015benchmark}
Nex F, Gerke M, Remondino F, Przybilla HJ, B{\"a}umker M, Zurhorst A (2015)
  {ISPRS} benchmark for multi-platform photogrammetry. ISPRS Annals of the
  Photogrammetry, Remote Sensing and Spatial Information Sciences
  II-3/W4:135--142

\bibitem[{Nex et~al.(2024)Nex, Zhang, Remondino, Farella, Qin, and
  Toschi}]{usegeo2024}
Nex F, Zhang N, Remondino F, Farella EM, Qin R, Toschi I (2024) {UseGeo} -- a
  {UAV}-based multi-sensor dataset for geospatial research. ISPRS Open Journal
  of Photogrammetry and Remote Sensing 12:100070

\bibitem[{{OpenDroneMap}(2016)}]{odmbellus}
{OpenDroneMap} (2016) odm\_data\_bellus.
  \url{https://github.com/OpenDroneMap/odm_data_bellus}, cC0 1.0, accessed
  2026-10-08

\bibitem[{{OpenDroneMap}(2022)}]{odmhelenenschacht}
{OpenDroneMap} (2022) odm\_data\_helenenschacht: Autel evo ii pro rtk survey
  with ground control.
  \url{https://github.com/OpenDroneMap/odm_data_helenenschacht}, gPL-3.0,
  accessed 2026-10-08

\bibitem[{Sarlin et~al.(2019)Sarlin, Cadena, Siegwart, and
  Dymczyk}]{sarlin2019hfnet}
Sarlin PE, Cadena C, Siegwart R, Dymczyk M (2019) From coarse to fine: Robust
  hierarchical localization at large scale. In: IEEE Conf. on Computer Vision
  and Pattern Recognition (CVPR)

\bibitem[{Sattler et~al.(2017)Sattler, Torii, Sivic, Pollefeys, Taira, Okutomi,
  and Pajdla}]{sattler2017largescale}
Sattler T, Torii A, Sivic J, Pollefeys M, Taira H, Okutomi M, Pajdla T (2017)
  Are large-scale 3d models really necessary for accurate visual localization?
  In: CVPR

\bibitem[{Sattler et~al.(2018)Sattler, Maddern, Toft, Torii, Hammarstrand,
  Stenborg, Safari, Okutomi, Pollefeys, Sivic, Kahl, and
  Pajdla}]{sattler2018benchmarking}
Sattler T, Maddern W, Toft C, Torii A, Hammarstrand L, Stenborg E, Safari D,
  Okutomi M, Pollefeys M, Sivic J, Kahl F, Pajdla T (2018) Benchmarking {6DOF}
  outdoor visual localization in changing conditions. In: IEEE Conf. on
  Computer Vision and Pattern Recognition (CVPR)

\bibitem[{Sch{\"o}nberger and Frahm(2016)}]{colmap}
Sch{\"o}nberger JL, Frahm JM (2016) Structure-from-motion revisited. In: IEEE
  Conf. on Computer Vision and Pattern Recognition (CVPR)

\bibitem[{Sch{\"o}ps et~al.(2017)Sch{\"o}ps, Sch{\"o}nberger, Galliani,
  Sattler, Schindler, Pollefeys, and Geiger}]{eth3d}
Sch{\"o}ps T, Sch{\"o}nberger JL, Galliani S, Sattler T, Schindler K, Pollefeys
  M, Geiger A (2017) A multi-view stereo benchmark with high-resolution images
  and multi-camera videos. In: IEEE Conf. on Computer Vision and Pattern
  Recognition (CVPR)

\bibitem[{Sch{\"u}{\ss}ler et~al.(2024)Sch{\"u}{\ss}ler, Fuchs, Sch{\"u}tze,
  Kuhn, Gunkel, Torizin, Pr{\"u}fer, and Tiepolt}]{wustrow2024}
Sch{\"u}{\ss}ler N, Fuchs M, Sch{\"u}tze K, Kuhn D, Gunkel C, Torizin J,
  Pr{\"u}fer S, Tiepolt L (2024) Wustrow bunker. Zenodo dataset,
  \doi{10.5281/zenodo.14002501}

\bibitem[{Sch{\"u}{\ss}ler et~al.(2025{\natexlab{a}})Sch{\"u}{\ss}ler, Fuchs,
  Sch{\"u}tze, Kuhn, Gunkel, Torizin, Pr{\"u}fer, and Tiepolt}]{sellin2025}
Sch{\"u}{\ss}ler N, Fuchs M, Sch{\"u}tze K, Kuhn D, Gunkel C, Torizin J,
  Pr{\"u}fer S, Tiepolt L (2025{\natexlab{a}}) Sellin sand cliff. Zenodo
  dataset, \doi{10.5281/zenodo.14655548}

\bibitem[{Sch{\"u}{\ss}ler et~al.(2025{\natexlab{b}})Sch{\"u}{\ss}ler, Fuchs,
  Sch{\"u}tze, Kuhn, Gunkel, Torizin, Pr{\"u}fer, and
  Tiepolt}]{wustrowerosion2025}
Sch{\"u}{\ss}ler N, Fuchs M, Sch{\"u}tze K, Kuhn D, Gunkel C, Torizin J,
  Pr{\"u}fer S, Tiepolt L (2025{\natexlab{b}}) Wustrow coastal erosion. Zenodo
  dataset, \doi{10.5281/zenodo.14655290}

\bibitem[{Torr(1997)}]{torr1997gric}
Torr PHS (1997) An assessment of information criteria for motion model
  selection. In: IEEE Conf. on Computer Vision and Pattern Recognition (CVPR)

\bibitem[{Triggs et~al.(2000)Triggs, McLauchlan, Hartley, and
  Fitzgibbon}]{triggs2000ba}
Triggs B, McLauchlan PF, Hartley RI, Fitzgibbon AW (2000) Bundle adjustment ---
  a modern synthesis. In: Vision Algorithms: Theory and Practice (ICCV
  Workshop), Springer, Lecture Notes in Computer Science, vol 1883, pp 298--372

\bibitem[{Umeyama(1991)}]{umeyama}
Umeyama S (1991) Least-squares estimation of transformation parameters between
  two point patterns. IEEE Trans on Pattern Analysis and Machine Intelligence
  13(4):376--380

\bibitem[{Wang(2019)}]{tuniu2019}
Wang YH (2019) Tuniu river, toufeng, miaoli county, taiwan (2019-04-11 and
  2019-09-16). OpenDroneMap community forum,
  \url{https://community.opendronemap.org/t/3292} and
  \url{https://community.opendronemap.org/t/3294}, accessed 2026-10-08

\bibitem[{{Wingtra AG}(2026)}]{wingtra}
{Wingtra AG} (2026) {WingtraOne} sample data sets.
  \url{https://wingtra.com/data-sets/}, surveys Turnpike 540 WSP and Malaysia
  urban cadaster

\end{thebibliography}

\end{document}